\documentclass[10pt,twocolumn,letterpaper]{article}
\PassOptionsToPackage{table}{xcolor}
\usepackage{cvpr}
\usepackage[T1]{fontenc}
\usepackage{newunicodechar}
\newunicodechar{–}{\textendash}

\usepackage{amsmath,amsfonts,bm}

\def\eqref#1{equation~\ref{#1}}

\def\1{\bm{1}}

\DeclareMathAlphabet{\mathsfit}{\encodingdefault}{\sfdefault}{m}{sl}
\SetMathAlphabet{\mathsfit}{bold}{\encodingdefault}{\sfdefault}{bx}{n}

\usepackage{makecell}
\usepackage{array}
\usepackage{multirow}
\usepackage{float}
\usepackage{placeins}
\usepackage{tikz}
\usetikzlibrary{arrows.meta,positioning,fit,calc,backgrounds}
\definecolor{batblue}{RGB}{230,242,255}
\definecolor{cvprblue}{rgb}{0.21,0.49,0.74}
\usepackage[pagebackref,breaklinks,colorlinks,allcolors=cvprblue]{hyperref}
\hypersetup{hypertexnames=false}
\def\confName{CVPR}
\def\confYear{2027}

\newcommand{\tablefootnotefont}{\fontsize{8}{9.2}\selectfont}

\title{BAT-WAM: Bridging Action and Touch in World-Action Models for Dexterous Hands}
\author{%
  \parbox{\dimexpr\textwidth-2\tabcolsep\relax}{%
    \centering\normalfont
    \textbf{Dong Fang}$^{1,2,*}$ \quad
    \textbf{Feixuan Huang}$^{2,3,*}$ \quad
    \textbf{Yuanxin Zhong}$^{2,\ddagger}$ \quad
    \textbf{Yuxiang Gao}$^{2,3}$ \\[0.3em]
    \textbf{Qi Song}$^{4}$ \quad
    \textbf{Yunlong Wang}$^{2}$ \quad
    \textbf{Xiaosong Jia}$^{1,\dagger}$  \\[0.8em]
    $^{1}$Fudan University \quad
    $^{2}$Rimbot \quad
    $^{3}$Southeast University \quad
    $^{4}$Tsinghua University
  }%
}
\begin{document}
\maketitle
\makeatletter
\iftoggle{cvprfinal}{%
\begingroup
\renewcommand{\thefootnote}{}
\renewcommand{\@makefntext}[1]{\parindent=0pt\noindent #1}
\footnotetext[0]{%
This work was conducted during an internship at Rimbot.\\
$^{*}$ These authors contributed equally to this work.\\
$^{\dagger}$ Corresponding authors.\\
$^{\ddagger}$ Project leader.
}
\endgroup
}{}
\makeatother
\begin{abstract}
Dexterous manipulation with world action models (WAMs) relies on hand configuration and contact information, yet joint-state and whole-hand tactile observations may be partially or entirely missing. RGB images provide scene context and hand–object geometry but are insufficient to reliably recover these missing modalities. We further observe a correlation between hand state and tactile information. Motivated by this observation, we introduce BAT-WAM, a framework that bridges visual, joint-state, and tactile representations for WAMs. BAT-WAM learns separate codebooks and decoders for the tactile and action streams. An RGB-conditioned continuous latent bridge connects the two streams for bidirectional learning, while visual information serves as continuous context. Under the HTBench evaluation protocol, BAT-WAM reduces the root mean squared error (RMSE) of tactile reconstruction by 34.39\% under fully masked conditions and 38.84\% under partially masked conditions, relative to an image-conditioned HandTouch checkpoint. For future tactile prediction conditioned on target-time RGB and historical actions, it reduces mean RMSE by 25.59\% relative to OpenWAM with HandTouch. Real-robot experiments further show that BAT-WAM consistently outperforms related WAM baselines in dexterous manipulation. Generalization experiments on unseen objects and scenes also demonstrate substantial improvements over baseline models. These findings support joint modeling of visual context, hand state, and touch as a representation approach for multimodal WAMs.
\end{abstract}

\section{Introduction}
\label{sec:introduction}

Contact-rich dexterous manipulation depends on the interplay between scene geometry, hand configuration, and physical contact. RGB images provide object appearance and visible hand--object geometry; tactile observations (Touch) capture contact locations and pressure distributions; and joint states (States) describe hand configuration and kinematic coordination. These modalities provide complementary information, but similar visual observations can conceal different interaction states. For example, two images of a hand grasping a cup may appear nearly identical, even when one grasp maintains stable contact and the other is beginning to slip. Such ambiguity motivates representations that capture fine differences in hand configuration and contact, consistent with the importance of tactile feedback in reactive manipulation \citep{niu2026trex}.
\begin{figure*}[t]
  \centering
  \includegraphics[width=\textwidth]{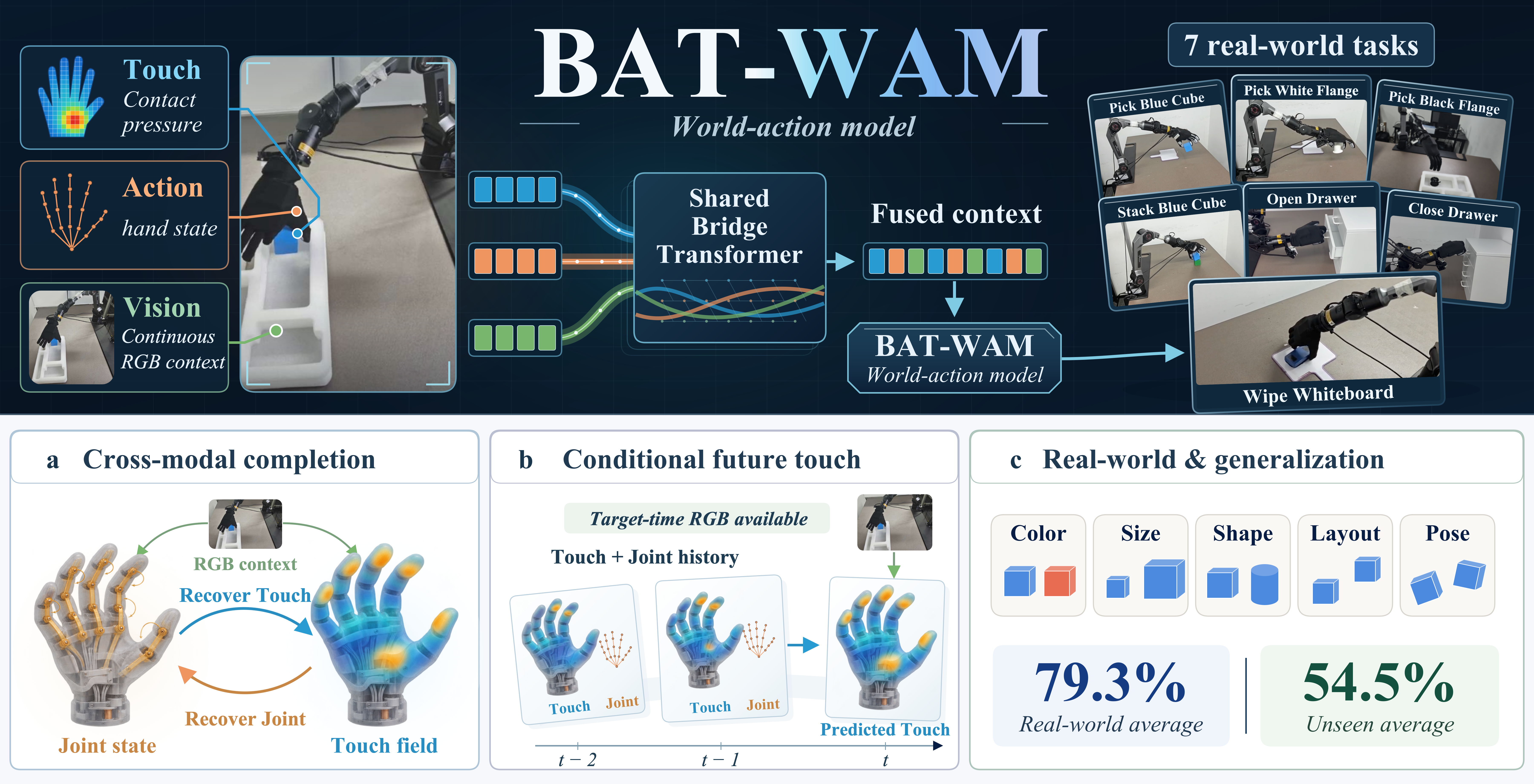}
  \caption{Overview. BAT-WAM bridges state and contact information with RGB observation.}
  \label{fig:architecture}
\end{figure*}

Recent work has developed several interfaces for incorporating proprioception and dexterous hand states into world action model (WAM) systems. For example, OpenWAM-$\alpha$~\citep{wang2026openwam} linearly
projects the robot's 80-dimensional proprioceptive state into a single context token. This token is appended to the language context and conditions both the video and action streams through cross-attention. For dexterous hands, UniDexTok learns discrete state representations through a standardized hand interface, while UniHM couples cross-embodiment hand-motion tokenization with language-guided manipulation \citep{fang2026unidextok,zhang2026unihm}. These approaches address the challenge of making structured, high-dimensional hand information accessible to token-based models.

In parallel, tactile VLA and world action model (WAM) research has expanded the role of touch in policy learning. Tactile-VLA and VLA-Touch incorporate tactile information into action generation, feedback, and control \citep{huang2025tactilevla,bi2025vlatouch}. VT-WAM, Tactile-WAM, and DreamTacVLA model future tactile observations alongside visual or action information \citep{tian2026vtwam,wu2026tactilewam,ye2025dreamtacvla}, while $N_0$-VTLA uses predictive latent tactile tokens to condition action generation \citep{neoteai2026n0vtla}. Building on these interfaces, we investigate a specific representation problem: how can Touch and Joint supervise one another within an RGB-conditioned interface while retaining their distinct physical meanings and data structures?

This question is motivated by the physical coupling between hand configuration and contact. Joint states constrain where fingers can engage an object, while tactile measurements provide evidence about the contacts realized by that configuration. RGB supplies the scene context needed to interpret these relationships. HT-Bench, which pairs egocentric vision with full-hand tactile observations, provides a basis for studying visually conditioned contact representations \citep{huang2026htbench}. We hypothesize that explicitly learning conditional dependencies between Touch and Joint can help resolve ambiguities that remain in visual observations and support inference when part of either modality is unavailable.

We introduce BAT-WAM, a multimodal representation interface designed for use by WAM backbones. BAT-WAM bridges state and contact information, while retaining RGB as the context. A continuous bridging module balances modality-specific token budgets and models interactions among hand configuration, contact state, and scene context. This design preserves distinct quantization and reconstruction pathways for Touch and Joint while allowing their representations to interact. Through conditional reconstruction, each modality can provide information for recovering the other: a modality serves as observed context in one masking configuration and as a supervision target in another, with RGB conditioning the inference.

The interface supports both static masked completion and conditional future tactile prediction. Each task explicitly specifies which observations are visible to the model and which signals are used only as supervision. Static completion examines reconstruction of missing information from the available context, while future prediction examines how observed joint-state history informs subsequent tactile states.

Our contributions are threefold:

\begin{enumerate}
    \item We uncover a previously overlooked cross-modal relationship: a strong correlation between dexterous hand states and tactile observations.

    \item Building on this finding, we introduce BAT-WAM, a novel cross-modal representation framework that integrates discrete tactile and state representations with continuous RGB context through a multimodal bridging mechanism.

    \item BAT-WAM achieves state-of-the-art performance on benchmarks for masked tactile completion and future tactile prediction, and outperforms the corresponding baselines in real-world experiments and zero-shot generalization evaluations.
\end{enumerate}
\begin{figure*}[t]
  \centering
  \includegraphics[width=\textwidth]{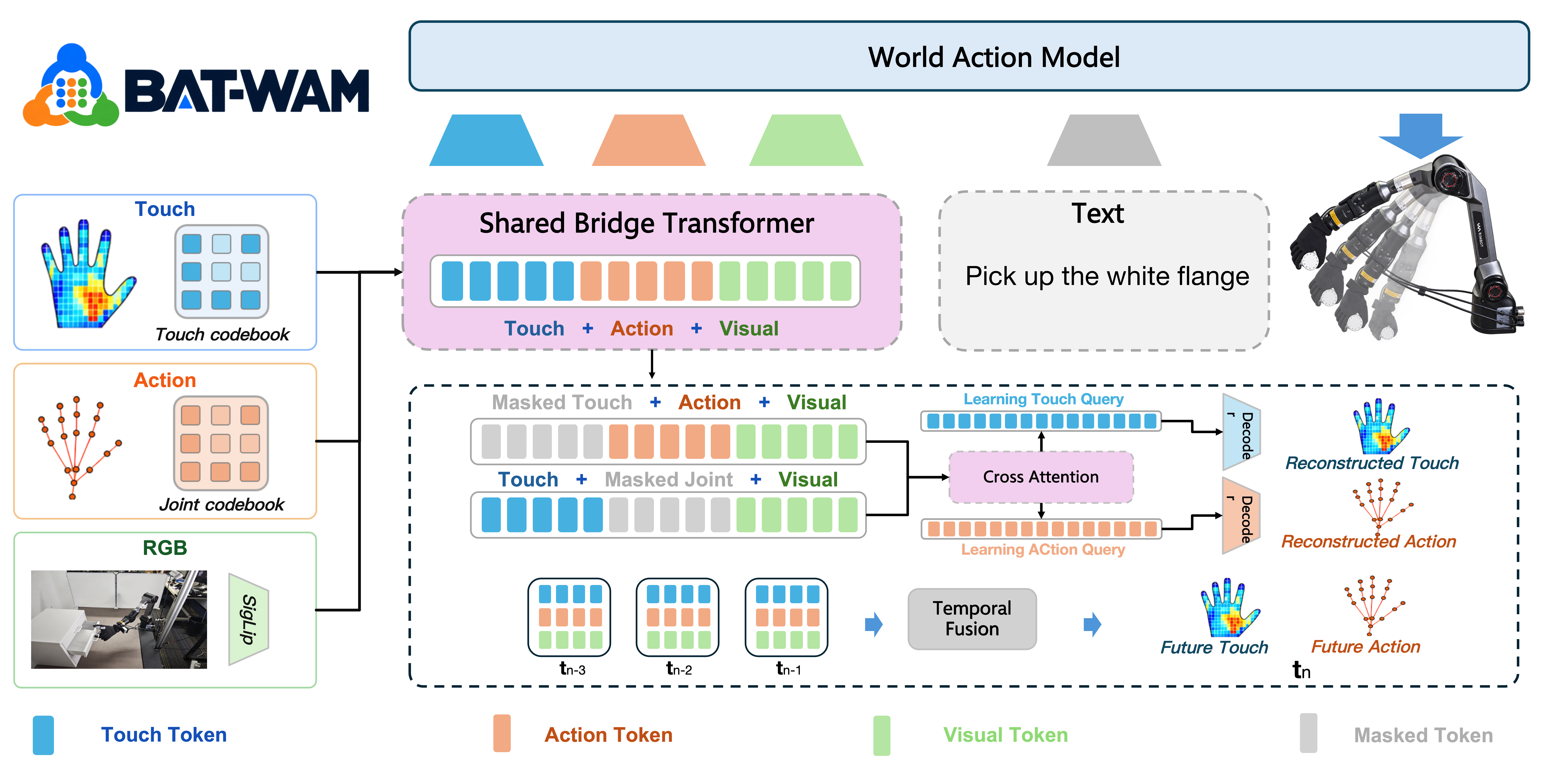}
  \caption{Pipeline of BAT-WAM.
A shared bridge Transformer integrates separately tokenized touch
and joint states with continuous RGB features.
Masked cross-modal reconstruction and temporal prediction train
the representation interface, whose outputs, together with language
instructions, condition a WAM backbone for action generation.}
  \label{fig:backbone_optimization}
\end{figure*}

\section{Related Work}
\label{sec:related_work}

\subsection{Proprioceptive State Representations}

Proprioceptive state representation learning focuses on encoding internal robot states, such as joint configurations, and integrating them with visual observations.
Existing methods adopt different interfaces for incorporating these states:
$\pi_0$ linearly projects continuous proprioceptive states into the Transformer embedding space~\citep{black2024pi0};
HPT uses embodiment-specific input encoders to map proprioceptive states and visual observations from heterogeneous robots into token sequences processed by a shared backbone~\citep{wang2024hpt};
and $\pi_{0.5}$ discretizes proprioceptive states and encodes them as text tokens through the language tokenizer~\citep{physicalintelligence2025pi05}.
These approaches explore continuous embeddings, heterogeneous input alignment, and discrete text encoding, providing different representation interfaces through which policy models can access proprioceptive information.
Proprioceptive states can also serve as reconstruction targets for representation learning.
For high-dimensional dexterous hands, UniDexTok addresses state heterogeneity across hand embodiments by organizing human and robot joint states through a unified 22-DoF semantic interface and learning embodiment-conditioned discrete state representations~\citep{fang2026unidextok}.
These studies provide foundations for state integration, cross-modal reconstruction, and unified state representations across embodiments.
Building on this direction, BAT-WAM incorporates standardized Joint states and Touch into a common representation interface while retaining separate codebooks and decoders for the two modalities.
Conditioned on continuous RGB context, it learns bidirectional dependencies between hand configuration and contact state through conditional reconstruction.

\subsection{Full-hand tactile representation}

Tactile representation learning includes continuous features, cross-sensor pretraining, and discrete tokenization. Sparsh learned visual-tactile representations with self-supervision, AnyTouch sought a common representation for static and dynamic information across visuotactile sensors, and VITaL supported tactile and non-tactile manipulation policies~\citep{higuera2024sparsh,feng2025anytouch,george2024vital}. VQ-VAE provided a standard route from continuous observations to a finite latent vocabulary~\citep{oord2017vqvae}, but pressure maps must also retain contact regions, spatial topology, and local intensity.

HTBench/HandTouch was the closest full-hand tactile representation setting to ours. It used bimanual pressure maps and egocentric vision to learn discrete tactile representations and reconstruct pressure maps under visual conditioning~\citep{huang2026htbench}. BAT-WAM retains this spatial pressure interface but gives Joint state its own codebook and decoder, so configuration information enters the bridge in a reconstructable form. 

\subsection{Tactile-enhanced VLA and world-action models}

Tactile information entered policies through continuous fusion, feedback control, prediction branches, and latent tokens. 3D-ViTac, Reactive Diffusion Policy, and T-Rex used vision and touch for fine-grained or reactive manipulation~\citep{huang2025_3dvitac,xue2025rdp,niu2026trex}, while TouchWorld developed a predictive model for tactile forecasting and reactive manipulation~\citep{zhou2026touchworld}. In VLA and WAM systems, Tactile-VLA and VLA-Touch used visual-language physical knowledge or layered feedback, while VT-WAM, Tactile-WAM, DreamTacVLA, and $N_0$-VTLA studied visuotactile action prediction, asymmetric attention, tactile world models, and latent tactile tokens~\citep{huang2025tactilevla,bi2025vlatouch,tian2026vtwam,wu2026tactilewam,ye2025dreamtacvla,neoteai2026n0vtla}.

These systems demonstrated the value of touch for contact control, but their inputs typically entered as continuous features, external feedback, prediction branches, or latent tokens. BAT-WAM specifies a narrower interface: Touch and states have separately decodable physical vocabularies, cross-modal relations are learned in a continuous bridge, RGB remains continuous context, and the resulting prefix is used for backbone observation optimization.

\section{Method}
\label{sec:method}

BAT-WAM encodes tactile observations and hand joint states under RGB conditioning,  Figure~\ref{fig:backbone_optimization} gives the complete interface.

\subsection{Modality-Specific Tokenizers}
\label{sec:tokenizers}

Touch is a dense spatial pressure field, whereas Joint is a low-dimensional vector describing the hand configuration. Given their different data structures, we use a separate encoder $E_m$, projection layer $P_m$, codebook $\mathcal{C}_m$, and physical decoder $D_m$ for each modality $m\in\{T,J\}$.

The encoder first extracts features, which the projection layer then maps into the quantization space. Let $u_m^i=P_m(E_m(x_m)^i)$ denote the $i$-th projected feature. Quantization is defined as
\begin{equation}
\begin{aligned}
    k_m^i
    &= \arg\min_k \left\|u_m^i-c_m^k\right\|_2^2, \\
    q_m^i
    &= u_m^i
    + \operatorname{sg}\!\left(c_m^{k_m^i}-u_m^i\right),
\end{aligned}
\label{eq:physical_quantization}
\end{equation}
where $c_m^k$ is the $k$-th codebook vector and $\operatorname{sg}$ denotes the stop-gradient operation. Touch and Joint use separate codebooks. Their quantized features are decoded into pressure maps and joint states by their respective physical decoders.

For each hand, Touch is encoded as 196 tokens on a regular spatial grid, and Joint is encoded as 8 tokens. RGB follows an independent continuous encoding path: a frozen SigLIP encoder extracts image patch features \citep{zhai2023siglip}, and learned queries produce context tokens from these features. RGB is neither codebook-quantized nor assigned a reconstruction objective.

\subsection{Cross-Modal Bridging}
\label{sec:cross_modal_bridge}

The cross-modal bridge integrates Touch, Joint, and RGB information. Touch describes contact locations and pressure distributions, Joint describes the current hand configuration, and RGB provides object and scene context. Each modality is processed through its own encoding path and mapped into a common 256-dimensional feature space, where a Transformer models their relationships.

Tokens receive the corresponding modality and view identity embeddings, allowing the model to distinguish their sources. RGB features remain continuous, and the RGB encoder remains frozen during training.

Taking Touch completion as an example, the model concatenates masked Touch features, visible states features, RGB features, and task-control information along the token dimension:
\begin{equation}
Z_{\mathrm{in}}
=
\operatorname{Concat}\!\left(
    \underbrace{Z_T^{\mathrm{mask}}}_{\text{masked Touch}},
    \underbrace{Z_J}_{\text{Joint}},
    \underbrace{Z_I}_{\text{RGB}},
    \underbrace{c_T}_{\text{task/mask cues}}
\right).
\label{eq:bridge_inputs}
\end{equation}
Here, $Z_T^{\mathrm{mask}}$, $Z_J$, and $Z_I$ denote the masked tactile, joint-state, and visual features, respectively. The control information $c_T$ specifies the task and masking configuration. The concatenated sequence is processed by the bridge Transformer:
\begin{equation}
H_T
=
\Phi\!\left(Z_{\mathrm{in}};V_T\right),
\label{eq:bridge_fusion}
\end{equation}
where $V_T$ specifies attention visibility and prevents access to hidden or unavailable information. The bridge output provides cross-modal features for constructing additional input tokens for the backbone.

The static training objective is
\begin{equation}
\begin{aligned}
\mathcal{L}_{\mathrm{static}}
={}&
\sum_{m\in\{T,J\}}
\left(
    \mathcal{L}_{\mathrm{anc}}^m
    +
    \mathcal{L}_{\mathrm{mask}}^m
\right) \\
&+
\lambda_{\mathrm{lat}}\mathcal{L}_{\mathrm{lat}}
+
\lambda_{\mathrm{code}}\mathcal{L}_{\mathrm{code}}
+
\lambda_{\mathrm{com}}\mathcal{L}_{\mathrm{com}}.
\end{aligned}
\label{eq:static_objective}
\end{equation}
Here, $\mathcal{L}_{\mathrm{anc}}^m$ supervises unimodal reconstruction, and $\mathcal{L}_{\mathrm{mask}}^m$ supervises cross-modal masked prediction. The latent alignment loss $\mathcal{L}_{\mathrm{lat}}$ aligns the bridge representation with the encoded features of the complete target. The terms $\mathcal{L}_{\mathrm{code}}$ and $\mathcal{L}_{\mathrm{com}}$ are the codebook matching and commitment losses, respectively.

\subsection{Multimodal Temporal Prediction}
\label{sec:temporal_prediction}

The static bridge learns the relationship between contact and hand configuration at the same time step. Its temporal extension uses historical observations to predict changes in contact. Let $t_0$ denote the start of the historical window and $t$ the target time. The model conditions on historical Touch observations $T_{t_0:t-1}$, historical Joint states $J_{t_0:t-1}$, and RGB observations $I_{t_0:t}$ that include the target time:
\begin{equation}
p_\theta\!\left(
    T_t
    \mid
    T_{t_0:t-1},\,
    J_{t_0:t-1},\,
    I_{t_0:t}
\right).
\label{eq:conditional_future_touch}
\end{equation}

With RGB at the target time available, this task uses historical contact and hand configuration to estimate the target Touch observation. Target-time Touch is used only for supervision. Joint inputs contain only historical observations and exclude the target-time state.
\begin{table*}[t]
\centering
\caption{Full-mask touch completion.}
\label{tab:full_mask_synthesis}
\normalsize
\begingroup
\definecolor{inputblue}{HTML}{2E86AB}
\definecolor{muted}{HTML}{777F84}
\providecommand{\tableoneinputyes}{\textcolor{inputblue}{\ensuremath{\checkmark}}}
\providecommand{\tableoneinputno}{\textcolor{muted}{\textendash}}
\setlength{\tabcolsep}{6pt}
\renewcommand{\arraystretch}{1.4}
\begin{tabular*}{\linewidth}{@{\extracolsep{\fill}}lccccc@{}}
\toprule
 & \multicolumn{3}{c}{Available inputs} & & \\
\cmidrule(lr){2-4}
Model & Full touch & Image & Action & F-RMSE $\downarrow$ & F-cIoU $\uparrow$ \\
\midrule
HandTouch & \tableoneinputno & \tableoneinputyes & \tableoneinputno & 0.0221 & 0.8316 \\
\addlinespace[2pt]
\midrule[0.3pt]
\addlinespace[2pt]
BAT-WAM & \tableoneinputno & \tableoneinputyes & \tableoneinputno & 0.0163 & 0.8828 \\
BAT-WAM & \tableoneinputno & \tableoneinputno & \tableoneinputyes & 0.0159 & 0.8836 \\
BAT-WAM & \tableoneinputno & \tableoneinputyes & \tableoneinputyes & \textbf{0.0145} & \textbf{0.8933} \\
\bottomrule
\end{tabular*}
\par\vspace{5pt}
{\fontsize{7.5}{10}\selectfont
\tableoneinputyes\ input available; \tableoneinputno\ input unavailable.
Bold indicates the best result in each metric.\par}
\endgroup
\end{table*}

\definecolor{inputblue}{HTML}{2E86AB}
\definecolor{inputgray}{HTML}{777F84}

\newcommand{\inputyes}{%
  \textcolor{inputblue}{\ensuremath{\checkmark}}}
\newcommand{\inputno}{%
  \textcolor{inputgray}{\textendash}}

\begin{table*}[t]
  \centering
  \caption{Local-mask touch completion.}
  \label{tab:local_mask_completion}

  \begingroup
  \normalsize
  \renewcommand{\arraystretch}{1.35}
  \setlength{\tabcolsep}{6pt}

  \begin{tabular*}{\textwidth}{
    @{\extracolsep{\fill}} l c c c c c @{}
  }
    \toprule
      & \multicolumn{3}{c}{Available inputs} & & \\
    \cmidrule(lr){2-4}
    Model
      & Partial touch
      & Image
      & Action
      & H-RMSE $\downarrow$
      & H-cIoU $\uparrow$ \\
    \midrule

    HandTouch
      & \inputyes & \inputyes & \inputno
      & 0.0224 & 0.8211 \\

    \midrule[0.3pt]

    BAT-WAM
      & \inputyes & \inputyes & \inputno
      & 0.0146 & 0.8925 \\

    BAT-WAM
      & \inputyes & \inputno & \inputyes
      & 0.0143 & 0.8934 \\

    BAT-WAM
      & \inputyes & \inputyes & \inputyes
      & \textbf{0.0137} & \textbf{0.8984} \\

    \bottomrule
  \end{tabular*}

  \par\vspace{5pt}
  \begin{minipage}{\textwidth}
    \footnotesize
    \inputyes: input available;
    \inputno: input unavailable.
    Bold indicates the best result in each metric.
  \end{minipage}
  \endgroup
\end{table*}
\begin{figure*}[t]
  \centering
  \includegraphics[width=0.8\textwidth]{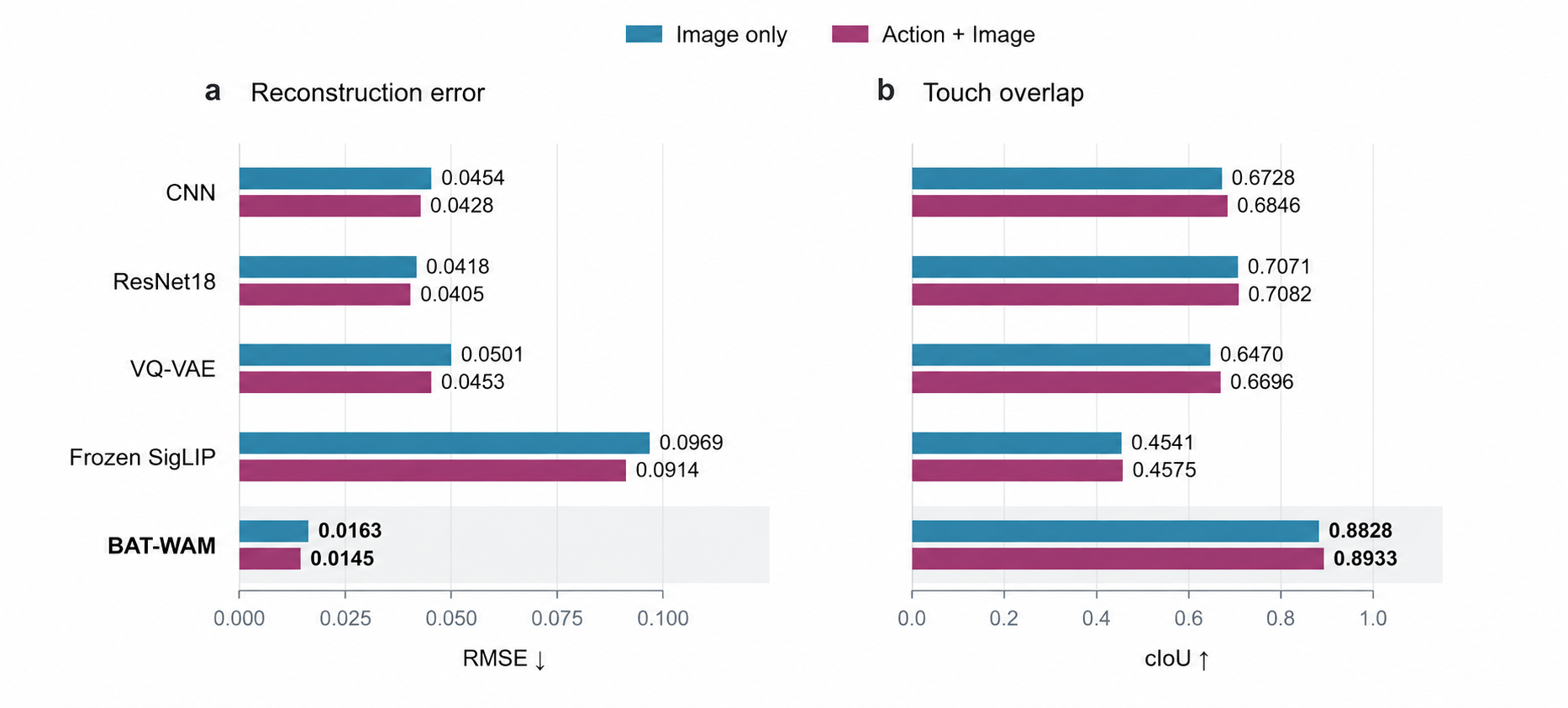}
  \caption{Under HTBench, BAT-WAM is the best on both displayed metrics for the Action+Image condition. }
  \label{fig:static_completion}
\end{figure*}

\section{Experiments}

\definecolor{inputblue}{HTML}{238BB5}
\definecolor{inputgray}{HTML}{808890}

\newcommand{\availcheck}{\textcolor{inputblue}{\ensuremath{\checkmark}}}
\newcommand{\unavaildash}{\textcolor{inputgray}{--}}
\begin{table*}[t]
    \centering
    \caption{Future tactile prediction.}
    \label{tab:temporal_prediction}

    \small
    \renewcommand{\arraystretch}{1.2}
    \setlength{\tabcolsep}{8pt}

    \begin{tabular*}{\textwidth}{
        @{\extracolsep{\fill}} l c c c c c @{}
    }
        \toprule
        & \multicolumn{3}{c}{Available inputs} & & \\
        \cmidrule(lr){2-4}
        Model
        & Touch
        & Image
        & Action
        & RMSE $\downarrow$
        & cIoU $\uparrow$ \\
        \midrule

        HandTouch
        & \availcheck
        & \availcheck
        & \unavaildash
        & 0.0211
        & 0.8486 \\

        \midrule

        BAT-WAM
        & \availcheck
        & \availcheck
        & \availcheck
        & \textbf{0.0157}
        & \textbf{0.8901} \\

        \bottomrule
    \end{tabular*}

    \vspace{4pt}
    \parbox{\textwidth}{
        \footnotesize
        \availcheck: input available;
        \unavaildash: input unavailable.
        Bold indicates the best result in each metric.\\
    }
\end{table*}

\label{sec:experiments}
 Our experiments comprise two main components: offline benchmark evaluation and real-world robot experiments. The offline evaluation addresses three questions: 
\begin{enumerate}   
    \item whether the continuous bridge can use Joint and RGB information when Touch is missing or partially masked;
    \item whether the same interface can predict future Touch from historical observations. 
    \item whether BAT-WAM outperform baseline models in real-world robotic experiments and demonstrate stronger generalization ability.
\end{enumerate}
Real-world experiments further demonstrate that BAT-WAM outperforms the baselines OpenWAM-$\alpha$ and OpenWAM-$\alpha$ + HandTouch.

\subsection{Experimental Setup }
\label{sec:experimental_setup}

\begin{figure*}[t]
    \centering
    \includegraphics[width=\textwidth]{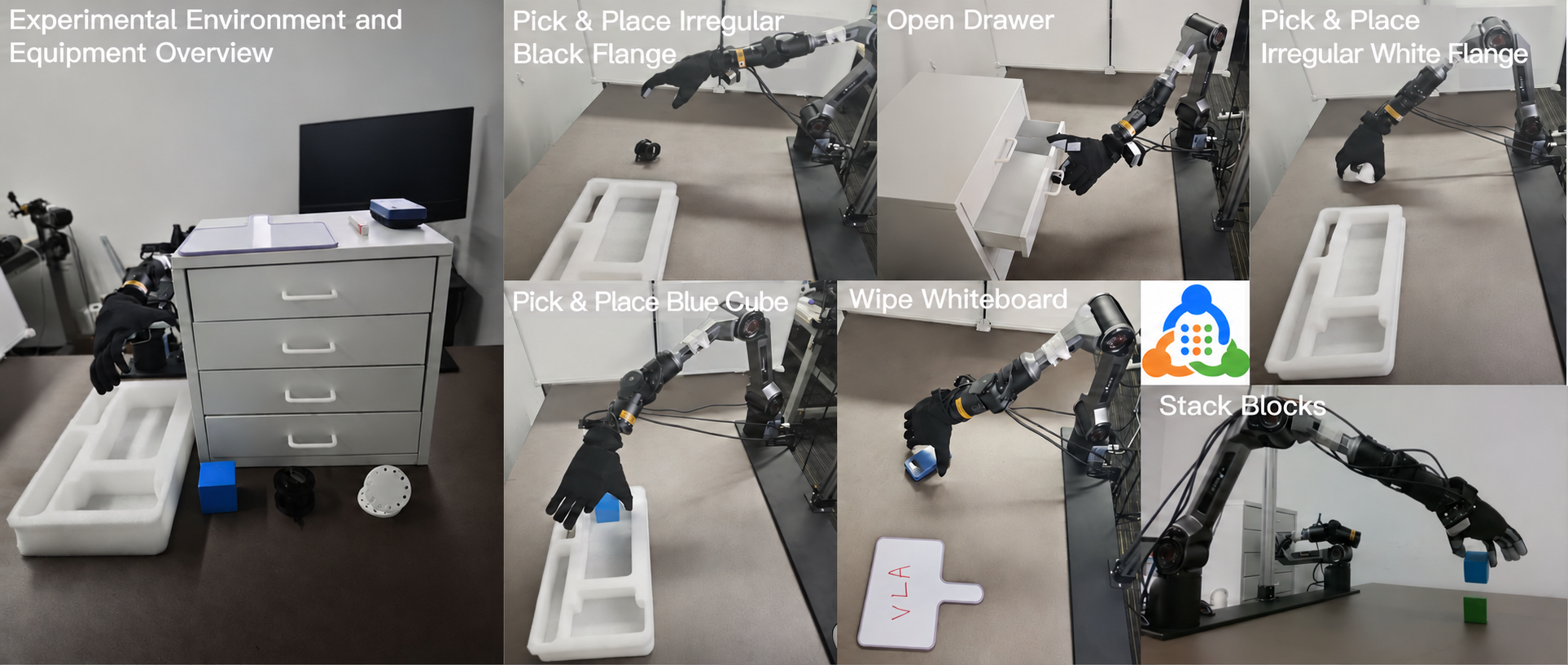}
    \caption{Real-world experiment environment and representative task examples. The left column shows the view of the shared tabletop setup.}
    \label{fig:real_world_setup_tasks}
\end{figure*}

\paragraph{Hardware.} 
We conduct real-world experiments using a 6-DoF AgileX PiPER robotic arm equipped with a Linker Hand L6 dexterous hand. The L6 hand is fitted with a full-hand tactile sensing glove from Juqiao Industrial. An Intel RealSense D435 provides top-view RGB-D observations, while
a wrist-mounted Intel RealSense D405 provides wrist-view RGB-D observations. The complete
experimental platform is shown in Figure~\ref{fig:real_world_setup_tasks}.

\paragraph{Tasks Description.}

We design 7 diverse tasks spanning pick\&place, stacking, pull\&push, and long horizon manipulation to evaluate our WAM models. For each task, we collect 200 demonstrations.

\paragraph{Benchmark.}
We use HT-Bench as our offline benchmark \citep{huang2026htbench}. HT-Bench is a multitask benchmark for tactile representation learning in robotic dexterous manipulation that combines egocentric vision with full-hand tactile data. HandTouch is the tactile representation learning method adopted in HT-Bench.
Through fine-grained tactile retrieval, masked tactile completion, vision-to-tactile synthesis, and multimodal tactile frame prediction, it evaluates contact structure modeling, visual--tactile alignment, temporal prediction, and generalization to unseen tasks.

Given ground-truth and predicted tactile values $T_u,\widehat{T}_u\in[0,1]$ at valid pixels $u\in\Omega$, we compute the evaluation metrics as follows:
\begin{equation}
\begin{aligned}
\operatorname{RMSE}(\Omega)
&=
\sqrt{
    \frac{
        \sum_{u\in\Omega}
        (\widehat{T}_u-T_u)^2
    }{
        |\Omega|
    }
}, \\
\operatorname{cIoU}(\Omega)
&=
\frac{
    \sum_{u\in\Omega}
    \min(\widehat{T}_u,T_u)
}{
    \max\!\left(
        10^{-12},
        \sum_{u\in\Omega}
        \max(\widehat{T}_u,T_u)
    \right)
}.
\end{aligned}
\label{eq:tactile_metrics}
\end{equation}

F-type metrics are computed over all valid tactile pixels, whereas H-type metrics are computed only over valid pixels hidden by the local mask. The cIoU metric measures the overlap between continuous tactile values, rather than the IoU of thresholded contact labels. For dataset-level evaluation, squared-error sums, pixel counts, and overlap statistics are accumulated across the entire evaluation set before computing the final metrics.

\begin{table*}[t]
    \centering
    \caption{Real-world success rates (\%) across different tasks.}
    \label{tab:real_world_results}

    \vspace{6pt}
    \renewcommand{\arraystretch}{1.5}

    \resizebox{\textwidth}{!}{%
        \begin{tabular}{@{}l|ccc|c|cc|c|c@{}}
            \toprule

            \multirow{3}{*}{\textbf{Method}}
            & \multicolumn{3}{c|}{\textbf{Pick \& Place}}
            & \textbf{Fine Manipulation}
            & \multicolumn{2}{c|}{\textbf{Pull \& Push}}
            & \textbf{Long Horizon}
            & \multirow{3}{*}{\textbf{Average}} \\

            \cmidrule(lr){2-4}
            \cmidrule(lr){5-5}
            \cmidrule(lr){6-7}
            \cmidrule(lr){8-8}

            & Pick
            & Pick
            & Pick
            & Stack
            & Open
            & Close
            & Wipe
            & \\

            & Blue Cube
            & White Flange
            & Black Flange
            & Blue Cube
            & Drawer
            & Drawer
            & Whiteboard
            & \\

            \midrule

            OpenWAM-$\alpha$
            & 80.0
            & 55.0
            & 60.0
            & 70.0
            & 85.0
            & \textbf{100.0}
            & 15.0
            & 66.4 \\

            OpenWAM-$\alpha$ + HandTouch
            & 75.0
            & 65.0
            & 65.0
            & 75.0
            & 85.0
            & 95.0
            & 25.0
            & 69.3 \\

            \rowcolor{batblue}
            BAT-WAM
            & \textbf{90.0}
            & \textbf{75.0}
            & \textbf{70.0}
            & \textbf{85.0}
            & \textbf{95.0}
            & 95.0
            & \textbf{45.0}
            & \textbf{79.3} \\

            \bottomrule
        \end{tabular}%
    }
\end{table*}
\begin{table*}[t]
    \centering
    \caption{Zero-shot success rates (\%) across different tasks.}
    \label{tab:zero_shot_results}

    \vspace{6pt}
    \renewcommand{\arraystretch}{1.5}

    \resizebox{\textwidth}{!}{%
        \begin{tabular}{@{}l|cc|cc|cc|cc|cc|c@{}}
            \toprule

            \multirow{3}{*}{\textbf{Method}}
            & \multicolumn{2}{c|}{\textbf{Color}}
            & \multicolumn{2}{c|}{\textbf{Size}}
            & \multicolumn{2}{c|}{\textbf{Shape}}
            & \multicolumn{2}{c|}{\textbf{Layout}}
            & \multicolumn{2}{c|}{\textbf{Pose}
            }
            & \multirow{3}{*}{\textbf{Unseen Avg.}} \\

            \cmidrule(lr){2-5}
            \cmidrule(lr){5-5}
            \cmidrule(lr){6-7}
            \cmidrule(lr){8-9}
            \cmidrule(lr){10-11}

            & Seen
            & Unseen
            & Seen
            & Unseen
            & Seen
            & Unseen
            & Seen
            & Unseen
            & Seen
            & Unseen\\

            \midrule

            OpenWAM-$\alpha$
            & 77.5
            & 67.5
            & 80.0
            & 20.0
            & 55.0
            & 10.0
            & 70.0
            & 57.5
            & 65.0
            & 37.5
            & 38.5\\

            OpenWAM-$\alpha$ + HandTouch
            & 75.0
            & 70.0
            & 80.0
            & 27.5
            & 60.0
            & 15.0
            & 67.5
            & 60.0
            & 70.0
            & 37.5
            & 42.0\\

            \rowcolor{batblue}
            BAT-WAM
            & \textbf{87.5}
            & \textbf{80.0}
            & \textbf{87.5}
            & \textbf{37.5}
            & \textbf{72.5}
            & \textbf{27.5}
            & \textbf{85.0}
            & \textbf{80.0}
            & \textbf{77.5}
            & \textbf{47.5}
            & \textbf{54.5} \\

            \bottomrule
        \end{tabular}%
    }
\end{table*}

\subsection{Multimodal Reconstruction and Temporal Prediction}

\label{sec:static_completion}
\label{sec:future_touch}
We evaluate BAT-WAM with three offline experiments. Full-mask reconstruction tests whether the modality-specific tokenizers retain touch and states signals. Local-mask reconstruction tests whether the continuous bridge can use Joint and RGB information when Touch is missing or partially masked. Conditional Future Touch prediction tests whether the same interface can predict future Touch from historical observations.

\paragraph{Full-mask reconstruction.}

Table~\ref{tab:full_mask_synthesis} compares image-conditioned HandTouch and BAT-WAM under the same sample identities, masks, and global-pixel evaluation protocol. BAT-WAM with Action+Image reaches F-RMSE 0.0145 and F-cIoU 0.8933, compared with 0.0221 and 0.8316 for HandTouch. This corresponds to a 34.73\% RMSE reduction; the absolute RMSE and cIoU improvements are 0.0077 and 0.0617, respectively. With Action and Image jointly available, BAT-WAM obtains the strongest full-mask Touch reconstruction, verifying that the modality-specific tokenizers retain and transmit touch and state signals through the shared interface.

\paragraph{Local-mask reconstruction.}

Table~\ref{tab:local_mask_completion} reports metrics on masked valid pixels. BAT-WAM with Action+Image reaches H-RMSE 0.0137 and H-cIoU 0.8984, compared with 0.0224 and 0.8211 for HandTouch. Hidden-region RMSE is reduced by 38.87\%; the absolute RMSE and cIoU improvements are 0.0087 and 0.0773, respectively. Mean and zero filling perform substantially worse, further showing that BAT-WAM learns structured completion from visible context.

When Touch is missing or partially masked, the continuous bridge jointly integrates the remaining Joint/Action and RGB context to recover the hidden region. Joint and RGB information therefore complement missing Touch and produce effective hidden-region reconstruction.

\paragraph{Conditional Future Touch prediction.}

Table~\ref{tab:temporal_prediction} compares conditional Future Touch prediction. In conditional future Touch prediction, BAT-WAM uses RGB, historical Touch observations, and historical actions to achieve an RMSE of 0.0157 and a cIoU of 0.8901. Compared with HandTouch, which uses RGB and historical Touch observations, BAT-WAM reduces RMSE by 25.59\% and increases cIoU by 4.15 percentage points, demonstrating better future Touch prediction performance under the evaluated input configurations.

Taken together, the full-mask experiment verifies that the modality-specific tokenizers retain Touch and Joint signals for reconstruction; the local-mask experiment verifies that the continuous bridge uses Joint and RGB context to complete missing Touch; and the Future Touch experiment verifies that the same interface predicts future Touch from historical observations. These results establish the effectiveness of the BAT-WAM shared interface for multimodal reconstruction, missing-modality completion, and temporal prediction.

\begin{figure}
    \centering
    \includegraphics[width=\linewidth]{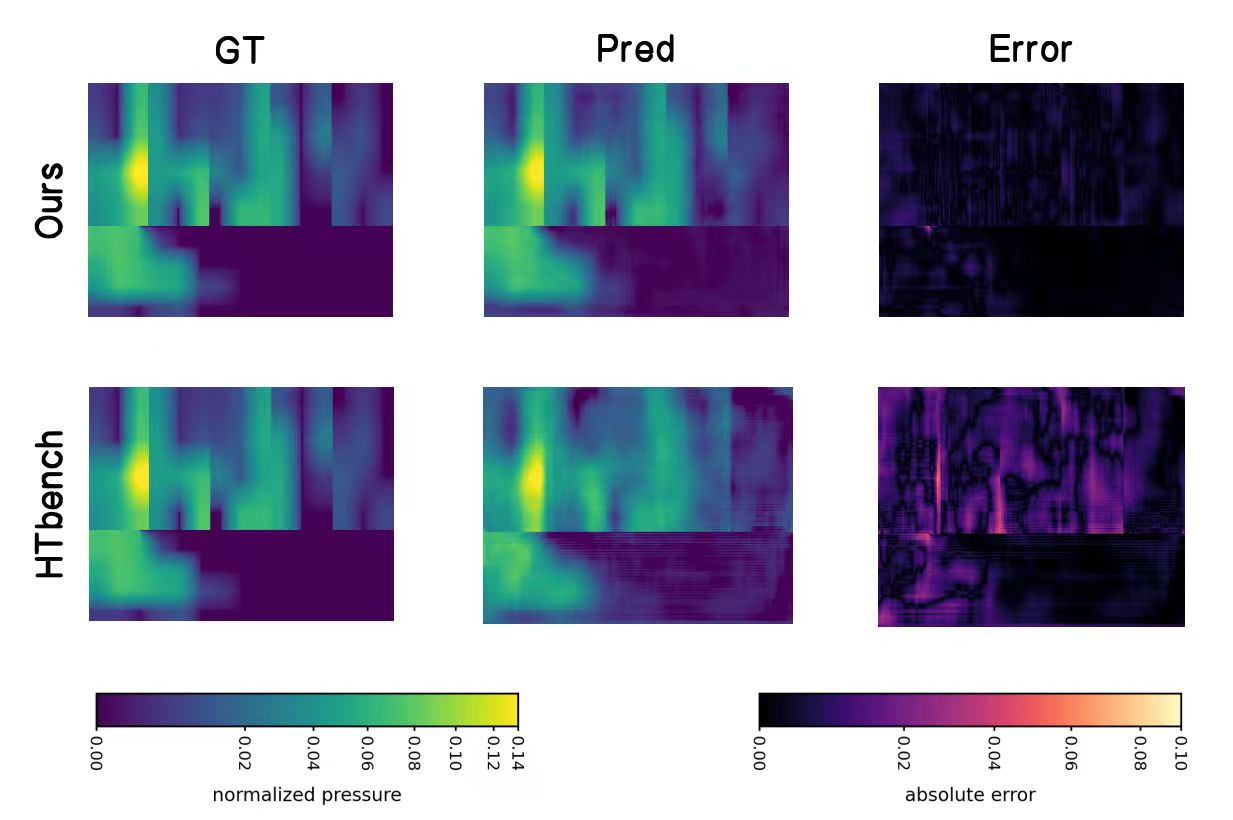}
    \caption{Tactile visualization comparison between Ht-Bench and our method.}
    \label{fig:placeholder}
\end{figure}

\subsection{Real-World Experiment}
\paragraph{Real World.}

Figure~\ref{fig:real_world_setup_tasks} presents seven real-world tasks spanning four categories---pick-and-place, fine manipulation, pull-and-push, and long-horizon tasks---all conducted in a shared tabletop setup. BAT-WAM achieves the highest mean success rate of 79.3\%, outperforming OpenWAM-$\alpha$ at 66.4\% and OpenWAM-$\alpha$ + HandTouch at 69.3\%. It achieves the highest success rate on every task except Close Drawer. The largest task-level advantage is observed on Wipe Whiteboard, where BAT-WAM achieves a success rate of 45\%, compared with 15\% for OpenWAM-$\alpha$ and 25\% for OpenWAM-$\alpha$ + HandTouch. We apply a strict success criterion to this task: an execution is considered successful only when all writing has been completely erased. This criterion helps explain the relatively low success rates.

Adding HandTouch to OpenWAM-$\alpha$ increases the mean success rate from 66.4\% to 69.3\% and improves performance on Pick White Flange, Pick Black Flange, Stack Blue Cube, and Wipe Whiteboard, suggesting that tactile information can benefit model performance. BAT-WAM further improves the mean success rate by 12.9 and 10.0 percentage points over OpenWAM-$\alpha$ and OpenWAM-$\alpha$ + HandTouch, respectively. These results demonstrate consistent transfer of BAT-WAM's multimodal interface across manipulation, grasping, wiping, and stacking tasks, yielding the strongest overall real-world performance among the evaluated methods.
\paragraph{Generalization.}
Table~\ref{tab:zero_shot_results} evaluates real-world zero-shot generalization across variations in object color, size, shape, scene layout, and object pose. BAT-WAM achieves the highest success rate in all five unseen settings, with an average of 54.5\%, compared with 38.5\% for OpenWAM-$\alpha$ and 42.0\% for OpenWAM-$\alpha$ + HandTouch, yielding gains of 16.0 and 12.5 percentage points, respectively. These improvements extend across multiple types of variation rather than being restricted to a single setting.

The largest absolute advantage occurs under unseen layouts, where BAT-WAM achieves 80.0\% success, compared with 57.5\% and 60.0\% for the two baselines, corresponding to improvements of 22.5 and 20.0 percentage points. BAT-WAM also reaches 80.0\% under unseen colors, demonstrating strong performance under the tested appearance and layout variations.

Generalization to unseen object sizes and shapes remains more challenging. BAT-WAM achieves success rates of 37.5\% and 27.5\%, respectively, substantially below the corresponding seen-setting results of 87.5\% and 72.5\%. Nevertheless, our design bridges action and tactile representations, and BAT-WAM achieves substantial improvements over OpenWAM-$\alpha$ in both challenging settings. Under unseen sizes, success increases from 20.0\% to 37.5\%, a gain of 17.5 percentage points and a relative improvement of 87.5\%. Under unseen shapes, success increases from 10.0\% to 27.5\%, also gaining 17.5 percentage points, with a relative improvement of 175.0\%.

These results suggest that bridging action and tactile representations can help mitigate the generalization challenges introduced by changes in object geometry. The equal absolute gains across size and shape variations show that the improvement is not limited to the larger relative gain observed under unseen shapes.



\section{Conclusion}
\label{sec:conclusion}

We introduced BAT-WAM, a multimodal representation interface that bridges tactile contact and hand configuration for world-action models. Separate codebooks and decoders encode tactile signals and hand states, while a shared Transformer learns their conditional relationships using continuous RGB context. This design supports static masked completion and conditional future tactile prediction within a common interface, connecting tactile observations with the hand configurations associated with them.

On HT-Bench, BAT-WAM improves full-mask tactile synthesis and local-mask completion under aligned evaluation protocols. It also improves conditional future tactile prediction using historical touch and states together with target-time RGB. Across seven real-world tasks in a shared tabletop setup, BAT-WAM achieves a mean success rate of 79.3\%, compared with 66.4\% for OpenWAM-$\alpha$ and 69.3\% for OpenWAM-$\alpha$ + HandTouch, corresponding to absolute gains of 12.9 and 10.0 percentage points, respectively. Beyond the standard task settings, BAT-WAM achieves an average zero-shot success rate of 54.5\% across five types of unseen variations, exceeding the two baselines by 16.0 and 12.5 percentage points.

\FloatBarrier
\section*{LLM Usage Statement}

OpenAI Codex was used to assist with adapting the LaTeX
source to the CVPR submission format, resolving compilation
issues, checking the layout, improving the clarity and
fluency of the manuscript's language.
{\small
\bibliographystyle{ieeenat_fullname}
\bibliography{references}
}
\clearpage
\setcounter{page}{1}
\maketitlesupplementary
\appendix

\setcounter{table}{0}
\setcounter{figure}{0}
\setcounter{equation}{0}
\renewcommand{\thetable}{S\arabic{table}}
\renewcommand{\thefigure}{S\arabic{figure}}
\renewcommand{\theequation}{S\arabic{equation}}

\section{Real-World Experimental Setup}
\begin{figure}
    \centering
    \includegraphics[width=1\linewidth]{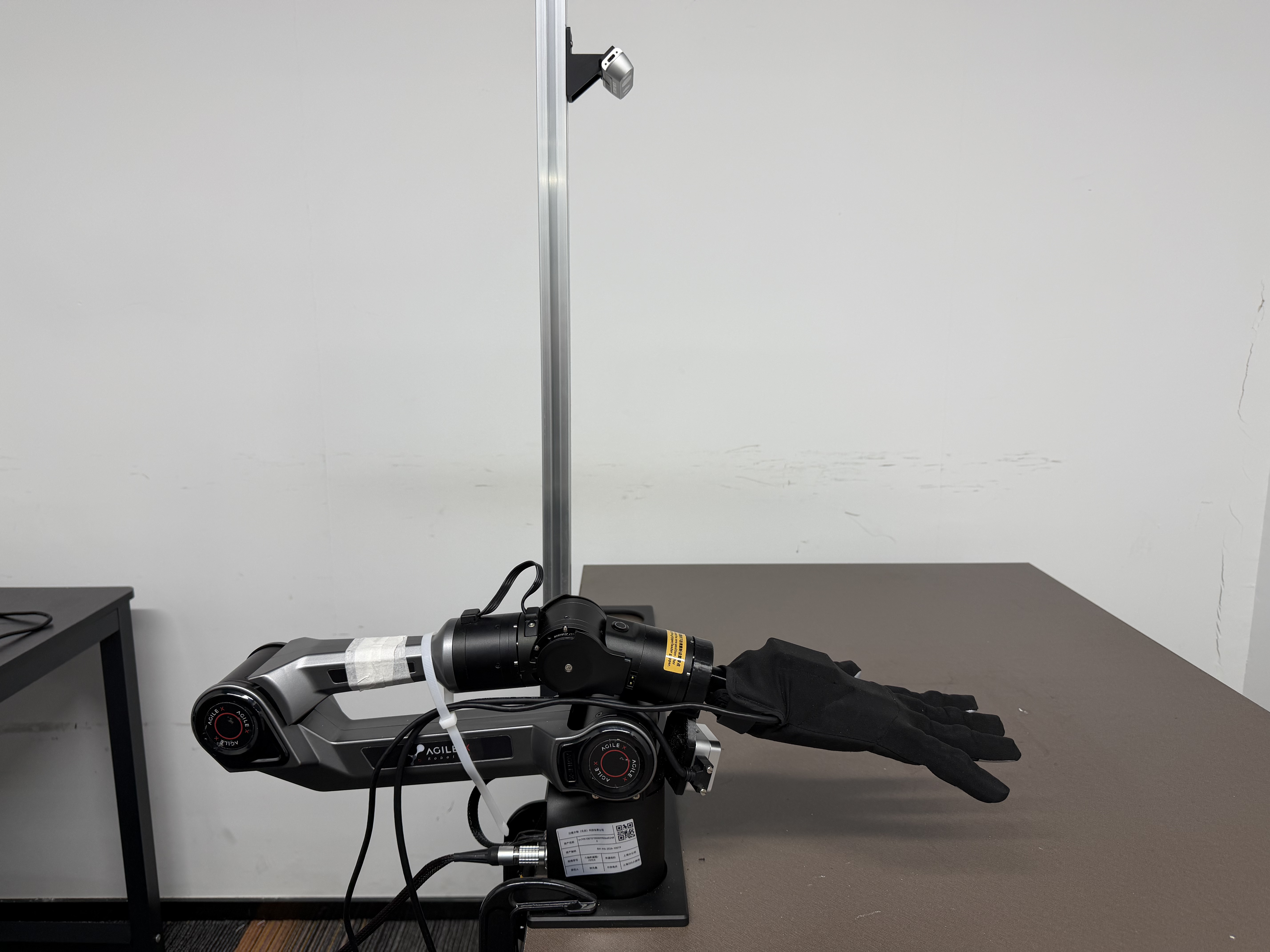}
    \caption{Side view of the experimental setup.}
    \label{fig:real_world_setup1}
\end{figure}
\begin{figure}
    \centering
    \includegraphics[width=1\linewidth]{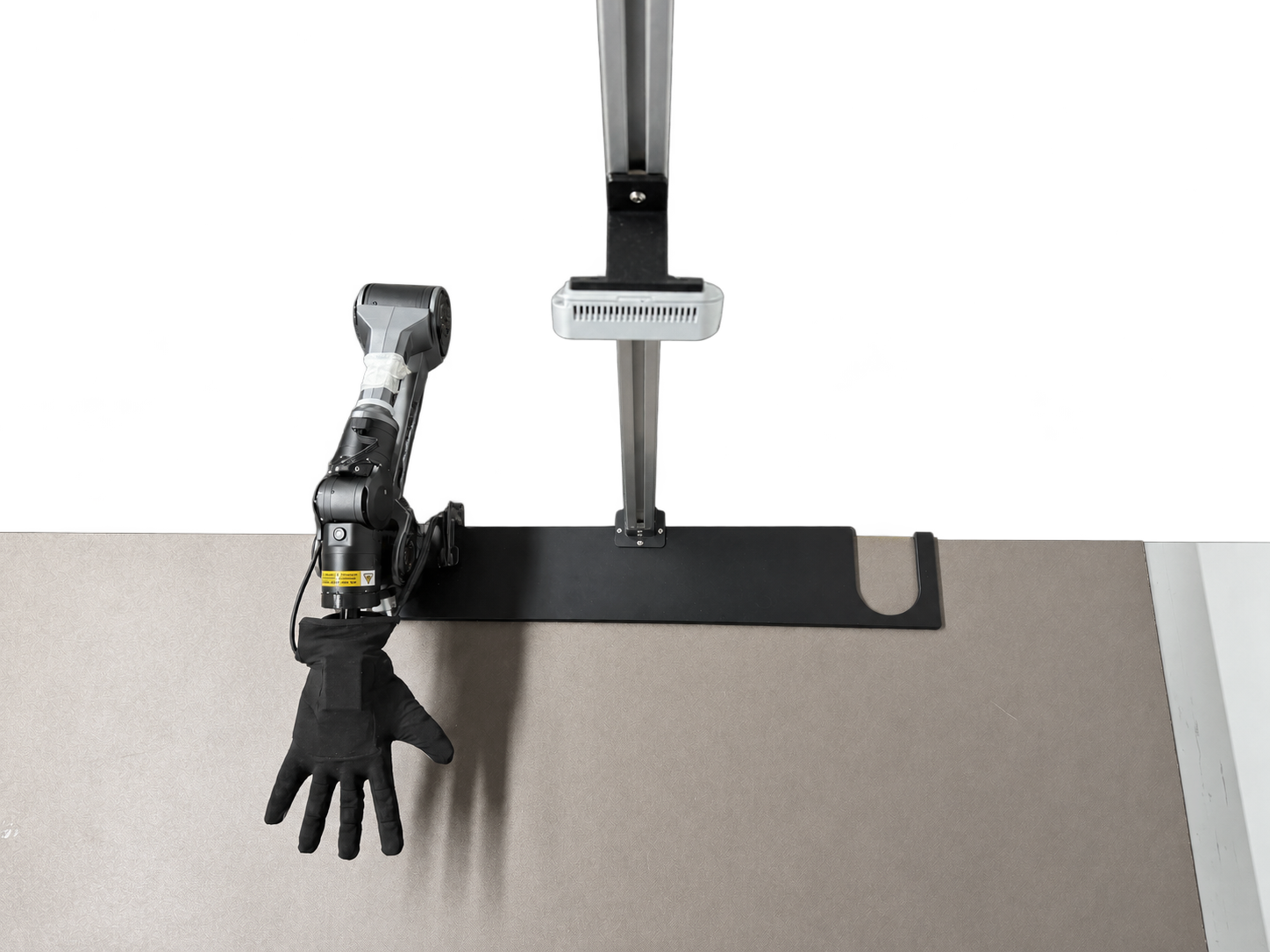}
    \caption{Front view of the experimental setup.}
    \label{fig:real_world_setup2}
\end{figure}
\paragraph{Tabletop setup.}
We conduct real-world tabletop experiments using a single
AgileX PiPER robotic arm equipped with different dexterous
hands across experiments. The arm is mounted at the edge
of the table, and dexterous hand is attached to its end effector through a corresponding 3D-printed flange adapter. 

In the main experiments, we use the Linker Hand L6. To further assess the robustness of our findings across tasks and hand configurations, we conduct additional real-world experiments that expand the set of tasks and introduce a second dexterous hand, the Wuji Hand, mounted on the same PiPER arm for comparison.
Figure~\ref{fig:real_world_setup1},~\ref{fig:real_world_setup2} show the tabletop setup with JUQIAO Tactile Glove.

\begin{figure}
    \centering
    \includegraphics[width=1\linewidth]{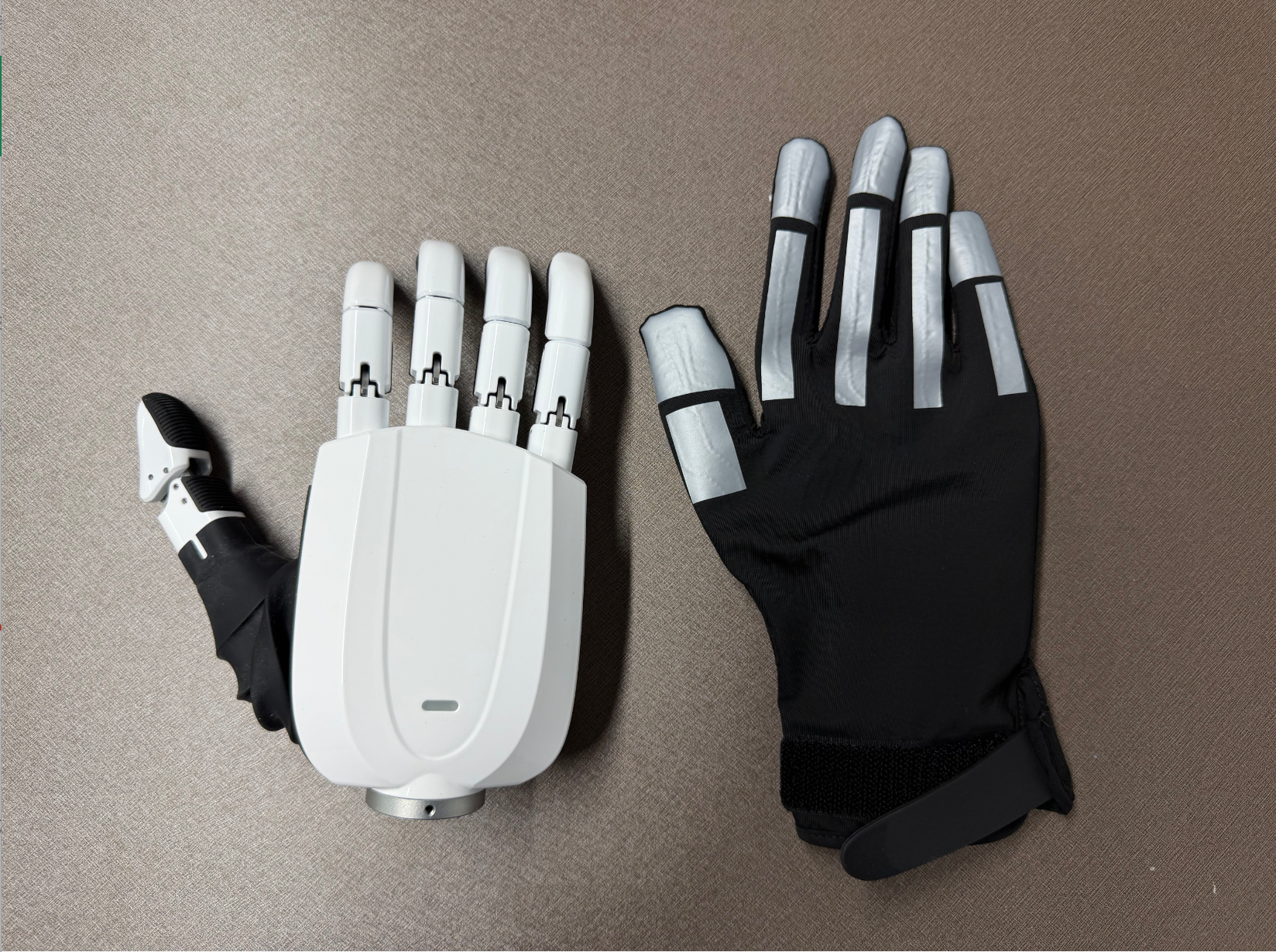}
    \caption{Dexterous hand and tactile glove.}
    \label{fig:hand}
\end{figure}
Figure~\ref{fig:hand} shows the LinkerHand L6 dexterous hand
(left) and the JUQIAO tactile glove (right) used in our experiments.

\paragraph{Camera setup.}
We use two Intel RealSense cameras to capture complementary
views of the manipulation scene. A RealSense D435 is mounted
on a fixed support beside the arm and provides top-view
RGB-D observations of the tabletop workspace. Its viewpoint
remains fixed as the arm moves, providing a common spatial
reference for observing object locations and their changes
throughout the task. A RealSense D405 is mounted beneath
the robot wrist and provides wrist-view RGB-D observations
near the dexterous hand. This camera moves with the end
effector and captures local details of the hand--object
interaction, including finger placement and the relative
positions of the hand and nearby objects. The two viewpoints
therefore provide complementary information at the workspace
and manipulation-region levels.

\definecolor{tablerowblue}{RGB}{229,242,255}
\begin{table*}[t]
    \centering
    \caption{Real-world success rates (\%) across different tasks.}
    \label{tab:additional_exp}
    \renewcommand{\arraystretch}{1.2}
    \resizebox{\textwidth}{!}{%
        \begin{tabular}{l|l|c|c|c|c}
            \toprule
            \textbf{Method}
            & \textbf{Hand}
            & \textbf{Vacuum Paper Scraps}
            & \textbf{Open Cup Lid}
            & \textbf{Pour Water}
            & \textbf{Average} \\
            \midrule

            \multirow{2}{*}{OpenWAM-$\alpha$}
            & Linker Hand L6
            & 30.0 & 80.0 & 60.0 & 56.7 \\
            & WUJI Hand
            & 20.0 & 90.0 & 30.0 & 46.7 \\
            \midrule

            \multirow{2}{*}{OpenWAM-$\alpha$ + HandTouch}
            & Linker Hand L6
            & 30.0 & 90.0 & 60.0 & 60.0 \\
            & WUJI Hand
            & 30.0 & 90.0 & 40.0 & 53.3 \\
            \midrule

            \rowcolor{tablerowblue}
            & Linker Hand L6
            & \textbf{50.0} & 90.0
            & \textbf{70.0} & \textbf{70.0} \\
            \rowcolor{tablerowblue}
            \multirow{-2}{*}{BAT-WAM}
            & WUJI Hand
            & 40.0 & \textbf{100.0}
            & 50.0 & 63.3 \\
            \bottomrule
        \end{tabular}%
    }
\end{table*}
\section{Additional Real-world Experiments}

We extend our real-world evaluation to 3 additional tasks using the PiPER robotic arm equipped with the Linker Hand L6 and the WUJI Hand.

As in the main experiments, we collect 200 trajectories for each task. We conduct 10 real-world evaluation trials per task and report the corresponding task success rates in Table~\ref{tab:additional_exp}.
BAT-WAM achieves the highest average success rate on both hands, reaching 70.0\% and 63.3\%, respectively, and outperforming OpenWAM-$\alpha$ + HandTouch by 10.0 percentage points in both settings. On Linker Hand L6, BAT-WAM improves success rates for vacuuming paper scraps and pouring water by 20.0 and 10.0 percentage points, respectively, while matching the baseline on opening a cup lid. On WUJI Hand, it improves all three tasks by 10.0 percentage points. These results support the effectiveness of BAT-WAM across the expanded task set and both hand configurations.

Although WUJI Hand exhibits lower average success rates across methods, the difference is task dependent: BAT-WAM achieves 100.0\% success on opening a cup lid with WUJI Hand, compared with 90.0\% with Linker Hand L6. A potential contributing factor to the lower average performance is the relatively small size of the JUQIAO tactile glove, which may result in a suboptimal fit on WUJI Hand and affect manipulation performance.

\section{Ablation Experiments}
\label{sec:app_ablations}
\begin{figure*}
    \centering
    \includegraphics[width=1\linewidth]{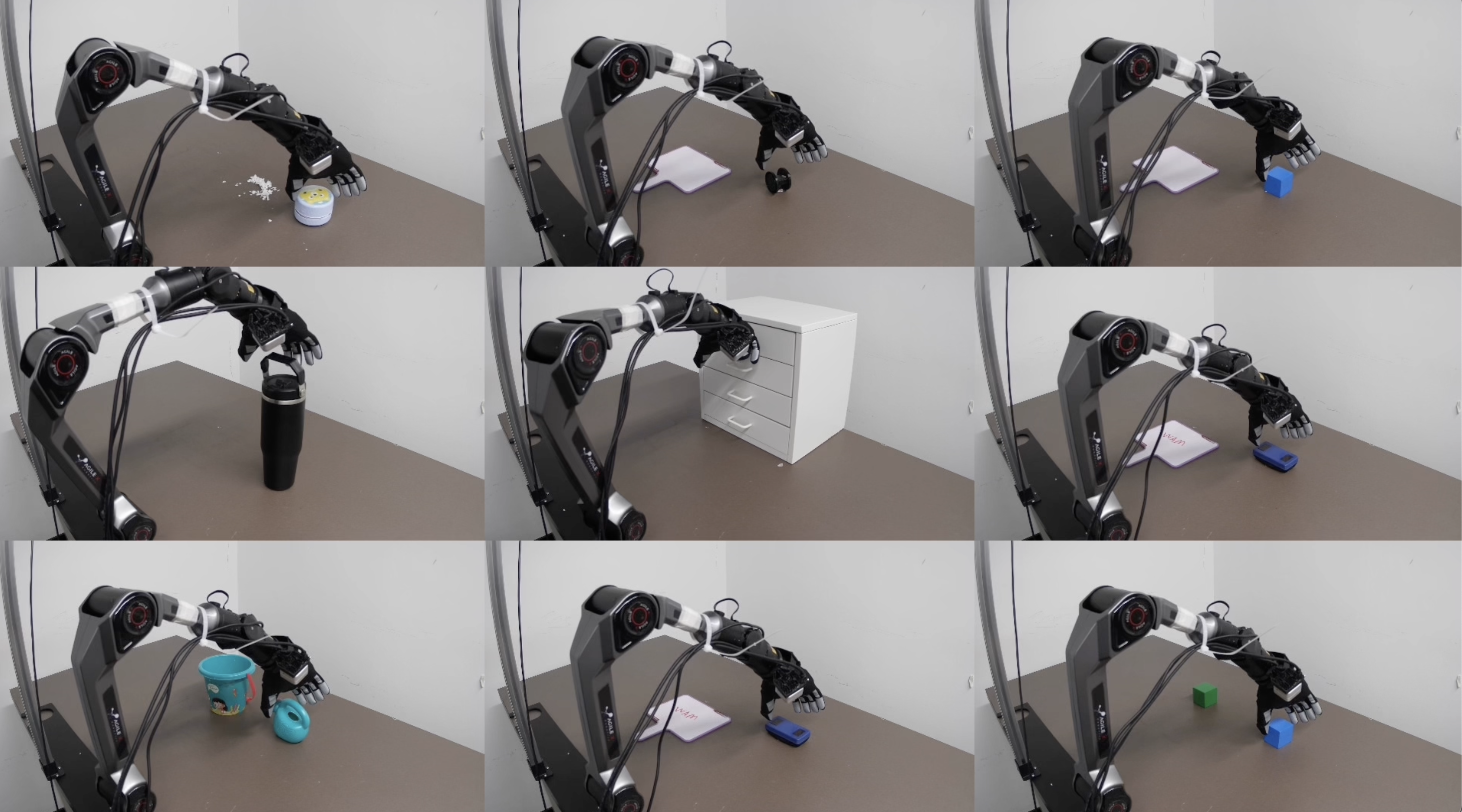}
    \caption{Task demonstrations. We design various tasks with tactile glove to test the model.}
    \label{fig:placeholder}
\end{figure*}
\begin{figure*}[!t]
\centering
\includegraphics[width=0.90\textwidth]{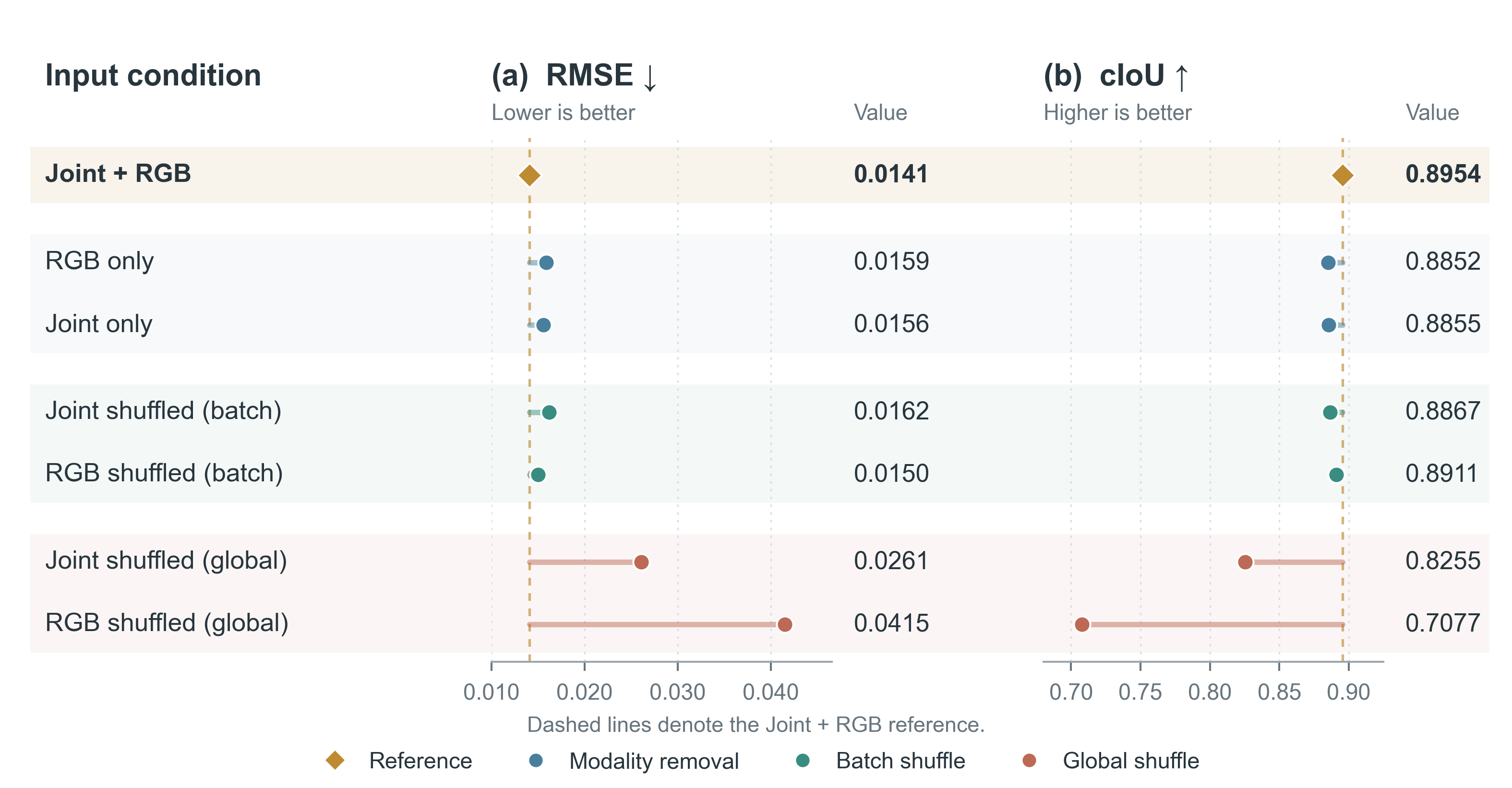}
\caption{Modality contribution and information controls. Left: RMSE;
right: cIoU. Global shuffles degrade substantially more than batch-local
shuffles, showing that the bridge uses sample-specific modality content rather
than a modality prior.}
\label{fig:app_modality}
\end{figure*}

\paragraph{Modality contribution.}
Table~\ref{tab:app_modality} and Figure~\ref{fig:app_modality} report ablations
that remove or corrupt one modality at test time. Removing Joint increases RMSE
by 12.27\%, and removing RGB increases it by 10.33\%, so both modalities are
necessary. A batch-local shuffle of Joint costs 14.40\%, i.e., \emph{wrong}
Joint is worse than \emph{missing} Joint; the bridge therefore uses the paired
joint content. Batch-local shuffling of RGB costs only 5.77\% because
neighbouring frames share scene and action context. A global shuffle across
scenes and episodes costs 193.36\%, confirming that the model relies on
sample-specific visual content.

\begin{figure*}
    \centering
    \includegraphics[width=0.9\linewidth]{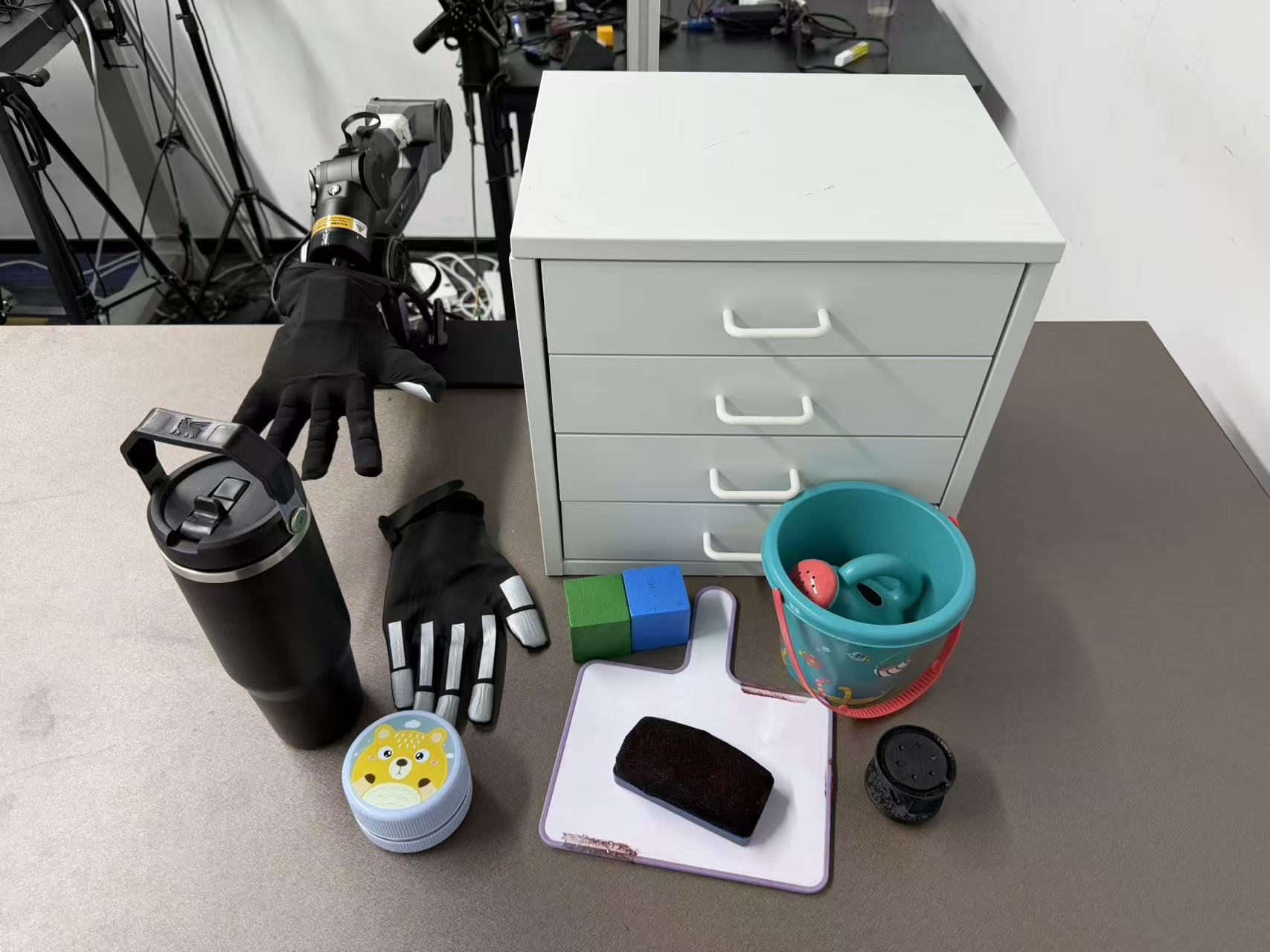}
    \caption{The complete set of objects and props used for real-world evaluation}
    \label{fig:placeholder}
\end{figure*}

\begin{table*}[!t]
\centering
\caption{Modality contribution and information controls on the ablation
subset. 
Global shuffles draw the modality from a different scene/episode; batch shuffles
draw from the same batch.}
\label{tab:app_modality}

\small
\renewcommand{\arraystretch}{1.15}
\setlength{\tabcolsep}{4pt}

\begin{tabular*}{\textwidth}{@{\extracolsep{\fill}}lcccc@{}}
\toprule
Setting & RMSE $\downarrow$ & cIoU $\uparrow$ & $\Delta$RMSE (\%) [95\% CI] & $\Delta$cIoU \\
\midrule
Joint + image (reference) & \textbf{0.0141} & \textbf{0.8954} & 0.00 & 0.0000 \\
Image only & 0.0159 & 0.8852 & +12.27 [+11.14, +13.45] & $-0.0101$ \\
Joint only & 0.0156 & 0.8855 & +10.33 [+9.09, +11.68] & $-0.0099$ \\
Batch-shuffled Joint & 0.0162 & 0.8867 & +14.40 [+13.06, +15.81] & $-0.0087$ \\
Batch-shuffled image & 0.0150 & 0.8911 & +5.77 [+5.33, +6.22] & $-0.0043$ \\
Globally shuffled Joint & 0.0261 & 0.8255 & +84.79 [+80.34, +89.36] & $-0.0699$ \\
Globally shuffled image & 0.0415 & 0.7077 & +193.36 [+184.39, +202.52] & $-0.1877$ \\
Left view duplicated & 0.0154 & 0.8863 & +8.79 [+7.86, +9.78] & $-0.0091$ \\
Center view duplicated & 0.0225 & 0.8372 & +59.09 [+52.41, +66.02] & $-0.0582$ \\
\bottomrule
\end{tabular*}

\par\vspace{2pt}
{\tablefootnotefont
Bold indicates the best result. ``duplicated'' repeats one view in both view
slots, so the model still sees two in-distribution RGB inputs.\par}
\end{table*}

\paragraph{View sensitivity.}
Duplicating the left view into both view slots costs 8.79\%, whereas duplicating
the center view costs 59.09\%. The left view is the primary view; the two-view
design remains preferable because the center view alone is insufficient.

\begin{figure}[H]
\centering
\includegraphics[width=0.90\linewidth]{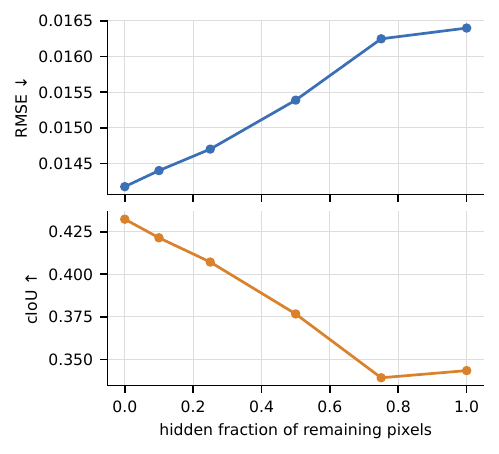}
\caption{Fixed-palm-hole context curve. The palm is always hidden; the
x-axis hides a growing fraction of the remaining sensor pixels. Error is
measured on the fixed palm hole. Visible Touch context reduces error by up to
13.6\%.}
\label{fig:app_context}
\end{figure}

\paragraph{Missing Touch.}
Figure~\ref{fig:app_context} evaluates the contribution of visible
tactile context by keeping the palm region masked and progressively
hiding additional sensor pixels outside the palm. Reconstruction
error is measured on the same palm region in every setting, allowing
us to examine how the availability of tactile observations elsewhere
affects prediction within the hidden region. As the visible Touch
context is removed, palm RMSE increases monotonically from 0.01417
to 0.01640. This progressive degradation indicates that the bridge
uses observed contact information from other regions of the hand
to reconstruct palm contact. Visible tactile observations therefore
provide information that complements the kinematic and visual cues
from Joint and RGB, rather than being ignored during completion.

\end{document}